\documentclass{acm_proc_article-sp}

\usepackage{url}
\usepackage{graphicx}
\usepackage{listings}
\usepackage{verbatim}
\usepackage{epstopdf}
\usepackage{algorithm2e}
\usepackage{rotating}
\usepackage{multicol}
\usepackage{multirow}
\usepackage{wrapfig}
\usepackage{color}
\usepackage{tikz}
\usetikzlibrary{arrows,shapes,positioning,calc,fit,backgrounds}
\usepackage{cite}
\usepackage{xspace}
\usepackage{longtable}
\usepackage{booktabs} 
\usepackage{array} 
\usepackage{marvosym}
\usepackage{stmaryrd}
\usepackage{etoolbox}
\usepackage[english]{babel}
\usepackage{hyperref}
\hypersetup{
    bookmarks=true, 
    pdftitle={Neural Programs}, 
    pdfauthor={Selyunin}, 
    pdfsubject={Neural Programs paper}, 
    pdfkeywords={TeX, LaTeX, graphics, images}, 
    colorlinks=false, 
	anchorcolor=black,
	runcolor=red,
    linkcolor=blue, 
    citecolor=black, 
    filecolor=black, 
    urlcolor=purple, 
	linkbordercolor=blue, 
	urlbordercolor=red, 
	citebordercolor=red, 
}

\lstdefinestyle{neural}{
  language=[ANSI] C,                 	   
  backgroundcolor=\color{white},   
  basicstyle=\scriptsize\ttfamily,        
  breakatwhitespace=false,         
  breaklines=true,                 
  captionpos=t,                    
  commentstyle=\itshape\color{mygray},    
  morekeywords={nwhile, nif},
  escapeinside={/**}{*/},          
  extendedchars=true,              
  frame=tb,                    
  keepspaces=true,                 
  keywordstyle=\bfseries\color{blue},       
  numbersep=5pt,                   
  numberstyle=\tiny\color{mygray}, 
  rulecolor=\color{black},         
  showspaces=false,                
  showstringspaces=false,          
  showtabs=false,                  
  stepnumber=1,                    
  stringstyle=\color{mymauve},     
  tabsize=2,                       
}

\definecolor{mygreen}{rgb}{0.133,0.545,0.133}
\definecolor{mygray}{rgb}{0.5,0.5,0.5}
\definecolor{mymauve}{rgb}{0.627,0.126,0.941}

\begin{document}

\title{
Deep Neural Programs for\\ 
Adaptive Control in Cyber-Physical Systems
}
%
%
%
%
%

\numberofauthors{1} 
%
\author{
%
%
\alignauthor
   K.~Selyunin, D.~Ratasich, E.~Bartocci, and R.~Grosu \\[2mm]
   \affaddr{Vienna University of Technlogy}\\
   \affaddr{konstantin.selyunin,denise.ratasich,ezio.bartocci,radu.grosu@tuwien.ac.at}
}
\date{30 September 2014}

\maketitle
\begin{abstract}
We introduce Deep Neural Programs (DNP), a novel programming paradigm
for writing adaptive controllers for cy\-ber-physical systems
(CPS). DNP replace if and while statements, whose discontinuity is
responsible for undecidability in CPS analysis, intractability in CPS
design, and frailness in CPS implementation, with their smooth, neural
nif and nwhile counterparts. This not only makes CPS analysis
decidable and CPS design tractable, but also allows to write robust
and adaptive CPS code. In DNP the connection between the sigmoidal
guards of the nif and nwhile statements has to be given as a Gaussian
Bayesian network, which reflects the partial knowledge, the CPS program
has about its environment. To the best of our knowledge, DNP are the
first approach linking neural networks to programs, in a way that
makes explicit the meaning of the network. In order to prove and
validate the usefulness of DNP, we use them to write and learn an
adaptive CPS controller for the parallel parking of the Pioneer rovers
available in our CPS lab.
\end{abstract}


\keywords{Cyber-Physical Systems, Adaptive Control, Car Parking,
  Deep Neural Pro\-grams, Gaussian Bayesian Networks.
} 

\vspace*{2mm}
\section{Introduction}
\label{sec:introduction}

Recent advances in sensing, actuation, communication, and
computation technologies, as well as innovations in their
integration within increasingly smaller, interconnected devices,
has lead to the emergence of a new and fascinating class of
systems, the so-called {\em cyber-physical systems
  (CPS)}. Examples of CPS include smart grids, smart factories,
smart transportation, and smart health-care~\cite{broyCPS}.

Similarly to living organisms, CPS operate in an uncertain,
continuously evolving ecosystem, where they compete for a limited
supply of resources. For survival, CPS need to continuously adapt,
such that, they react in real time and optimal fashion, with regard to
an overall survival metric, their partial knowledge, and their bounded
sensing, actuation, communication and computation capabilities.

\begin{figure}[htbp]
\vspace*{1mm}
\begin{center}
\includegraphics[width=\linewidth]{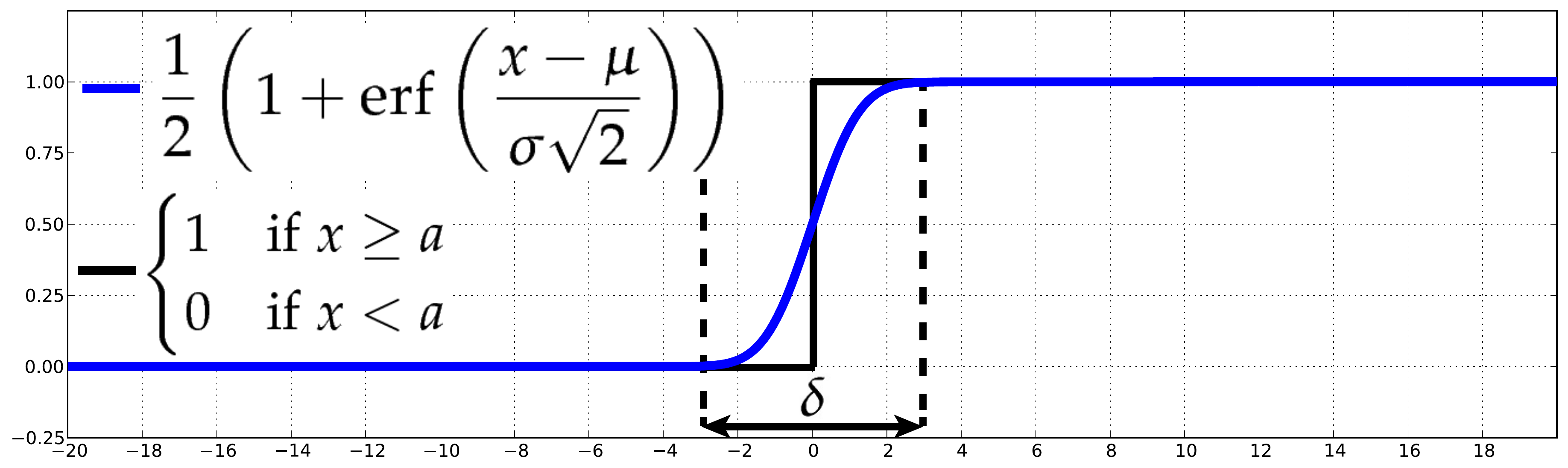}
\end{center}
\vspace*{-2mm}
\caption{Sigmoid (blue) and step (black) functions.}
\label{fig:neuron}
\end{figure}

In order to equip CPS with such exceptional features, various
researchers have started to wonder weather our current CPS
analysis, design and implementation techniques are still
adequate.  Going back to Parnas, Chaudhuri and Lezama identified
in a series of intriguing papers~\cite{Parnas85,Chaudhuri10,Chaudhuri11}, the
if-then-else construct as the main culprit for program frailness.
In a simple decision of the form \texttt{if\,(x\,{>}\,a)}, the
predicate $x\,{>}\,a$ acts like a step function (see
Figure~\ref{fig:neuron}), with infinite plateaus to the left and
to the right of the discontinuity point $x\,{=}\,a$. In a typical
mid-size program, the nesting of thousands of if-then-else
conditions leads to a highly nonlinear program, consisting of a
large number of plateaus separated by discontinuous jumps. This
has important implications.

From a CPS-analysis point of view, predicates of the form
$f(x)\,{>}\,a$, where $f(x)$ is a nonlinear analytic function, are a
di\-sas\-ter. They render \emph{CPS analysis undecidable}.
Intuitively, in order to separate all points on one side of the curve
$f(x)\,{=}\,a$, from all on the other side, one needs to forever
decrease the size of a grid, in all the rectangles that are crossed by
the curve. Such a process does never terminate, except for linear
functions where computation is still prohibitive. For this reason, a
series of papers, of Fraenzle, Ratschan, Wang, Gao and
Clarke~\cite{Fraenzle99,Ratschan06,Wang14,Gao14}, proposed the use of
an indifference region $\delta$ (see Figure~\ref{fig:neuron}), and
rewrite the predicates in the form $f(x){-}a\,{>}\,\delta$. This
approach not only makes program analysis (wrt.~reals) decidable, and
computable in polynomial time, but it also aligns it with
the finite computational precision available in today's computers.

From a CPS-design point of view, where one is interested to find the
values of $a$ for which an optimization criterion is satisfied,
predicates of the form $f(x)\,{>}\,a$ are a nightmare. They render
\emph{CPS optimization intractable}.  Intuitively, a gradient-descent
method searching for a local minimum, gets stuck in plateaus, where a
small perturbation to the left or to the right, still keeps the search
on the same plateau.  In order to alleviate this problem, Chaudhuri
and Lezama~\cite{Chaudhuri10} proposed to smoothen the steps, by
passing a Gaussian input distribution through the CPS. This can be
thought of as corresponding to the sensing and actuation noise. The
parameters of this distribution control the position of the resulting
sigmoidal curve (see Figure~\ref{fig:neuron}), and its steepness, that
is, the width of the above indifference region $\delta$. The authors
however, stopped short of proposing a new programming paradigm, and
the step-like functions in the programs to be optimized, posed
considerable challenges in the analysis, as they cut the Gaussians in
very undesirable ways.

From a CPS-implementation point of view, conditional statements of the
form \texttt{if\,(f(x)\,{>}\,a)} are also a disaster. They render
\emph{CPS frail and nonadaptive}. In other words, a small change in
the environment or the program itself, may lead to catastrophic consequences,
as the CPS is not able to adapt. In the AI community, where steps are
called \emph{hard neurons} and sigmoid curves are called \emph{soft
  neurons}, adaptation and robustness is achieved by learning a
particular form of Ba\-ye\-si\-an networks with soft-neuron distributions,
called neural networks. Such networks, and in particular deep neural
networks, have recently achieved amazing performance, for example in
the recognition of sophisticated
patterns~\cite{Ciresan12,Erhan10}. This technology looks so promising
that major companies such as Google and Amazon are actively recruiting
experts in this area. However, the neural-networks learned are still
met with considerable opposition, as it is very difficult, if not
impossible, to humanly understand them.

Having identified the if-then-else programming construct as the
major source of trouble in the analysis, design and
implementation of CPS, the following important question still
remains: \emph{Is there a simple, humanly understandable
  way to develop robust and adaptive CPS?} It is our belief, that
such a way not only exists, but it is also amazingly simple!

First, as program skeletons express domain knowledge and
developer intuition, they are here to stay. However, one needs to
replace hard neurons with their soft counterparts. We call such
program statements neural if-then-else statements, or
nif-then-else for short. They represent probabilistic, probit
distributions, and the decision to choose the left or the right
branch is sampled from their associated Gaussian
distributions. As a consequence, a program with nif statements
represents not only one, but a very large (up to the computational
precision) set of correct executions.

Second, the partial knowledge of such a program is encoded as a
Bayesian network, expressing the conditional dependencies among
the Gaussian distributions occurring within the nif
statements. These dependencies may be given, learned through a
preliminary phase and continuously improved during deployment, or
inferred through optimization techniques. In this case, learning
and optimization are considerably simplified, as the program is
by definition smooth. The depth and the branching structure of
these Bayesian networks reflect the sequential and parallel
nesting within the program, which is an essential asset in
program understanding.

For example, a parallel algorithm for pattern recognition, may
possess a quad-tree Bayesian structure, hierarchically reflecting
the neighbourhood relation among subimages. The depth of the
network is determined by the height of the tree. Similarly, a
purely sequential program, representing successive decisions,
will have a very linear Bayesian structure, whose depth is
determined by the number of decisions.

In order to validate our new paradigm, we use the parking example
from~\cite{LezamaSlides10}. The goal of this example was to
automatically learn the parameters of a program skeleton,
intuitively expressing the control as follows: Go backwards up to
a point $a_1$, turn up to an angle $b_1$, go backwards up to
$a_2$, turn again up to $b_2$ and finally go backwards up to
$a_3$. Since this program uses classical if statements, it is not
adaptive, and a small perturbation such as a slippery
environment, may lead to an accident. We therefore rewrite the
program with nif statements, and learn the conditional Gaussian
network associated with the predicates within these statements.
Using its sensors, the control program is now able to detect the
actual stopping or turning points, and to adequately sample its
next targets. Although this program is written once and for all,
it is able to adapt to a varying environment.

The main contributions of the work presented in this paper can be
therefore briefly summarized as follows: 
\vspace*{-3.5mm}
\begin{enumerate}
\item We propose a new programming paradigm for the development of
  smooth and adaptive CPS in which:
\vspace{-1mm}
\begin{itemize}
\item Troublesome ifs are replaced by neural ifs, thus improving
  analysis, design and implementation,
\item Partial knowledge is encoded within a learned Ba\-ye\-sian
  network, with Gaussian distributions.
\end{itemize}
\vspace*{-2mm}
\item We demonstrate the versatility of this programming paradigm
  on a parking example using Pioneer rovers. The associated
  youtube videos are available at~\cite{neuralVideos}.
\end{enumerate}
\vspace*{-4mm} Given obvious space limitations, we do not address
CPS-analysis and CPS-design (optimization) in this paper. They will be
the subject of a series of follow up papers.

The rest of the paper is organized as follows. In
Section~\ref{sec:background} we introduce Bayesian inference, Bayesian
networks, and Gaussian and Probit distributions.  In
Section~\ref{sec:neural} we introduce our programming paradigm. In
Section~\ref{sec:learning} we discuss how to learn the Gaussian
Bayesian network. In Section~\ref{sec:experiments} we discuss our
implementation platform and the associated results.  In
Section~\ref{sec:related} we discuss related work. Finally in
Section~\ref{sec:conclusion} we give our concluding remarks and
directions for future work.

\vspace*{2mm}
\section{Background}
\label{sec:background}


%



The main tool for logical inference is the \emph{Modus-Ponens} rule:
Assuming that proposition $A$ is true, and that from the truth of $A$
one can infer the truth of proposition $B$, one can conclude that
propositions $A$ and $B$ are both true. Formally:
\[
A \wedge (A \rightarrow B) ~=~ A \wedge B ~=~ B \wedge (B \rightarrow A)
\]
In probability theory, the uncertainty in the truth of a proposition
(also called an event) is expressed as a probability, and implication
between propositions is expressed as a conditional probability. This
leads to a probabilistic extension of Modus-Ponens, known as the
\emph{Bayes' rule}. Formally:
\[
P(A)~ P(B \mid A) ~=~ P(A \wedge B) ~=~ P(B)~ P(A \mid B)
\]
This rule, consistent with logic, is the main mechanism for
probabilistic inference~\cite{russellnorvig}. It allows to reason in
both forward, or causal way, and backwards, or diagnostic way. For
example if $B$ is causally implied by $A$, then the left term in the
above equation denotes a causal relation, and the right term, a
diagnostic relation. Equating the two, allows one to use causal
information (or observed events), for diagnostic inference.
In real-world systems, causal relations are usually chained and can
form sophisticated structures. 

\vspace*{-4mm}
\paragraph{Bayesian Networks}
A probabilistic system is completely characterized by the joint
probability distribution of all of its (possibly noisy)
components. However, the size of this distribution typically explodes,
and its use becomes intractable.  In such cases, the Bayes' rule,
allows to successively decompose the joint distribution according to
the conditional dependences among its \emph{random variables
  (RV)}. These are both discrete or continuous variables, which
associate to each value (or infinitesimal interval) in their range,
the rate of its occurrence. Networks of conditional dependencies
among random variables are known as \emph{Bayesian networks (BN)}, and
they have a very succinct representation.

Syntactically, a BN is a direct acyclic graph $G\,=\,(V,E)$, where
each vertex $v_i \in V$ represents a random variable $X_i$ and each
edge $e_{ij} \in E$ represents a conditional dependence of the
variable $X_j$ on the variable $X_i$. To avoid the complications
induced by the use of the joint probability distribution (or density),
each variable $X_i$ is associated with a \emph{conditional probability
  distribution (CPD)} that takes into account dependencies only
between the variable and its direct
parents~\cite{russellnorvig,Koller09}. Such a compact representation
keeps information about the system in a distributed manner and makes
reasoning tractable even for large number of variables.

Although in many interesting applications the variables of a BN have
discrete distributions (e.g.~in fault detection, a device might have
only a finite number of diagnosable errors, caused by a finite set of
faults), in many other applications, continuous random variables
naturally describe the entities of interest. For instance, in our
\emph{parallel parking} running example, a Pioneer rover starting from
an initial location, needs to execute a sequence of motion primitives
(e.g.~driving or turning forward or backward with fixed speed for a
particular distance $X_i$ or angle $X_j$), which will result in parking
the rover in a dedicated parking spot.

\vspace*{-4mm}
\paragraph{Gaussian Distributions}
Any real measurement of a physical quantity is affected by
noise. Hence, the distances and the angles occurring in our parking
example are naturally expressed as continuous RVs.  In this paper we
assume that variables have \emph{Normal}, also called,
\emph{Gaussian distributions (GD)}. These distributions naturally
occur from the mixing of several (possibly unobservable) RVs,
and they have mathematical properties making them very attractive.

An \emph{univariate Gaussian distribution (UGD)} is denoted by
$\mathcal{N}(\mu,\sigma^2)$ and it is characterized by two parameters:
The \emph{mean} $\mu$ and the \emph{variance} $\sigma^2$. In our
example, the desired distance in the first motion is associated with
$\mu$, which is perturbed by noise with variance $\sigma^2$. The
\emph{Gaussian probability density} of a RV $X$ with values $x$ is
defined as follows:
\begin{equation}
  \texttt{pdf}_{\mu, \sigma^2}(x) =
  \frac{1}{\sqrt{2\pi} \sigma}exp \left(
  -\frac{(x - \mu)^2}{2 \sigma^2} \right).
\end{equation}
Parallel parking includes a sequence of motion primitives that are
mutually dependent. To express these dependencies we use a
\emph{multivariate Gaussian Distribution (MGD)} \cite{grimmett_book},
which generalizes the Gaussian distribution to multiple dimensions.
For a $n$-dimensional vector of random variables $\mathbf{X}$ the
probability density function is characterized by a $n$-dimensional
mean vector $\mathbf{\mu}$ and a symmetric positive definite
covariance matrix $\mathbf{\Sigma}$.  To express the probability
density of a multivariate Gaussian distribution we use the inverse of
covariance matrix, called precision matrix
$\mathbf{T}=\mathbf{\Sigma}^{-1}$, which will be helpful later
during the learning phase. The probability density then can be written
as follows\cite{Neapolitan:2003}:
\begin{equation}
  \texttt{pdf}_{\mu, \sigma^2}(\mathbf{x}) =
  \frac{1}{(2\pi)^{n/2}(det(\mathbf{T}^{-1}))^{1/2}}exp \left(
  -\frac{1}{2} \Delta^2(\mathbf{x}) \right),
\end{equation}
where $\Delta^2(\mathbf{x}) = (\mathbf{x} -
\mathbf{\mu})^{T}\mathbf{T}(\mathbf{x} - \mathbf{\mu})$.

A \emph{Gaussian BN (GBN)} is a BN where random variables $X$
associated to each node in the network have associated a Gaussian
distribution, conditional on their parents $X_i$.

\vspace*{-4mm}
\paragraph{Probit Distributions}
In order to smoothen the decisions in a program, we need to choose a
function without plateaus and discontinuities.  Since we operate with
Gaussian random variables, the natural candidate is their
\emph{cumulative distribution function (CDF)}.  This is an
\emph{S}-shaped function or a \emph{sigmoid} (see
Figure~\ref{fig:nwhile_sampling}), whose steepness is defined by
$\sigma^2$, where $\operatorname{erf}$ denotes the error function:
\begin{equation}
  \texttt{cdf}_{\mu, \sigma^2}(x) = \frac12\left(1 +
  \operatorname{erf}\left( \frac{x-\mu}{\sigma\sqrt{2}}\right)\right),
  \label{}
\end{equation}
For a particular value $x$ of $X$, the function $\texttt{cdf}_{\mu,
  \sigma^2}(x)$ returns the probability that a random sample from the
distribution $\mathcal{N}(\mu, \sigma^2)$ will belong to the interval
$(-\infty,x]$.

Since the sensors and actuators of the Pioneer rover are noisy, the
trajectories it follows are each time different from the optimal one
(assuming that such difference is tolerated by the parking
configuration), even if the optimal trajectory of the parking example
is known. To be adaptive we use a combined approach: we incorporate
probabilistic control structures in the program (introduced in the
Section~\ref{sec:neural}) and sample commands from a GBN, whose
parameters were learned experimentally. To detect changes in the
environment and get more accurate position estimates, data from 
various sensors are combined with a sensor fusion algorithm.


\vspace*{2mm}
\section{Neural Code}
\label{sec:neural}

\begin{figure}[htbp]
\begin{center}
\includegraphics[width=0.9\linewidth]{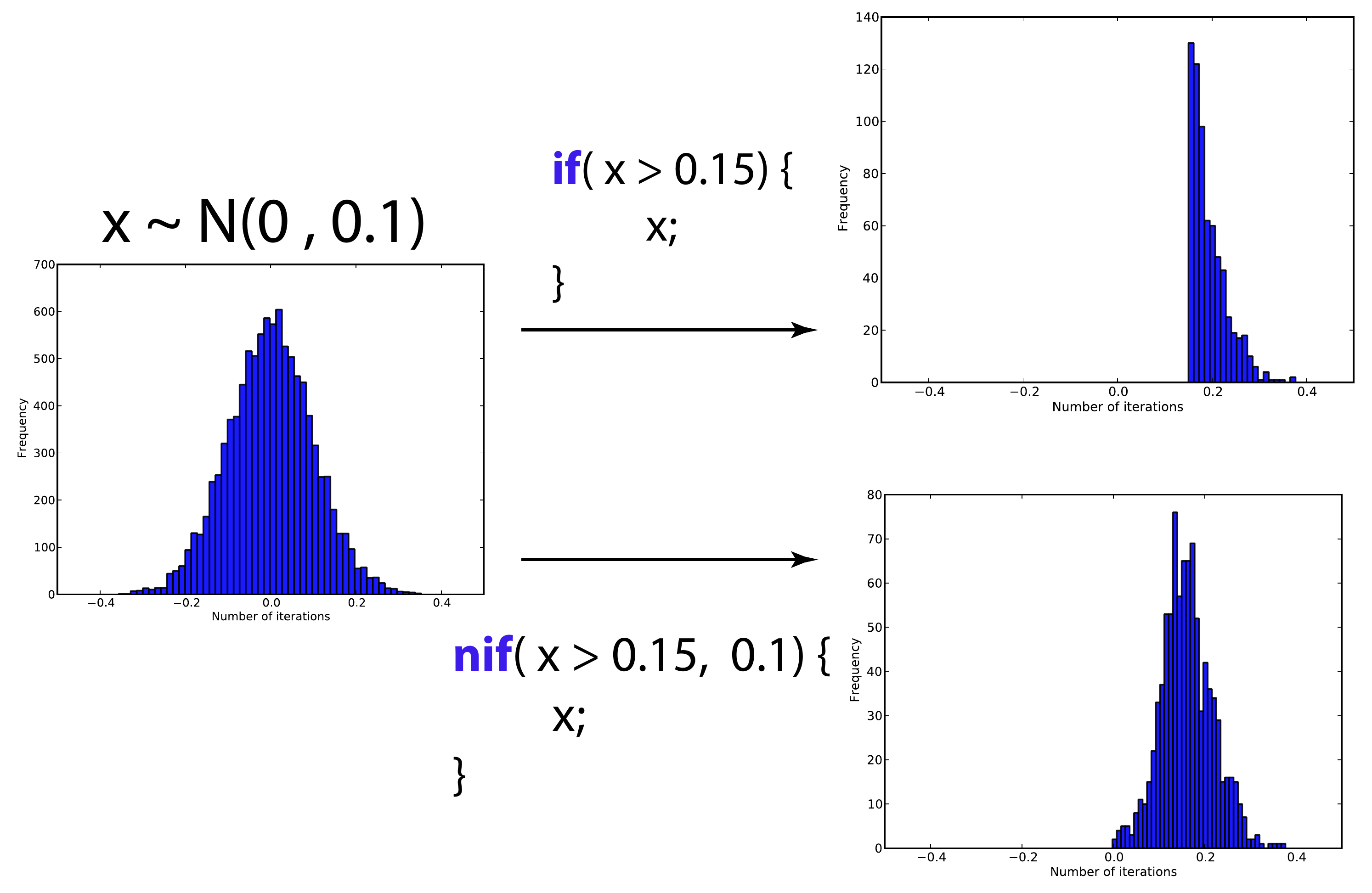}
\end{center}
\vspace*{-4mm}
\caption{Passing RVs through conditions}
\label{fig:nif_cut}
\end{figure}

Traditional inequality relations (e.g.~$>$, $\geq$, $\leq$, $<$)
define a sharp (or firm) boundary on the satisfaction of a condition,
which represents a step function (see Fig~\ref{fig:neuron}).  Using
firm decisions in a program operating on Normal RVs, cuts
distributions in halves, leaving unnormalized and invalid PDFs (see
Figure~\ref{fig:nif_cut}: The upper right plot shows the approximation
of the PDF after passing a Normal RV through a traditional conditional
statement).  Hence, to keep a proper PDF after passing a Normal RV
through an \texttt{if} or a \texttt{loop} statement one needs to
perform a re-normalization of the PDF.

In order to avoid re-shaping of probability density each time after a
variable is passed through a condition, we introduce a special type of
control structure called \emph{neural if}, or \texttt{nif} for short.
The name is coined to express the key novelty of our approach: We
propose to use a smooth conditionals $\texttt{cdf}_{\mu, \sigma^2}(x)$
instead of firm ones (see Figure~\ref{fig:neuron}).  A \texttt{nif}
statement operates on an inequality relation and a variance
$\sigma^2$, and decides which branch should be taken:
\texttt{nif(x~$\#$~y,$\sigma^2$)}, where $\#$ can be replaced with
($>$, $\geq$, $\leq$ or $<$) and $\sigma^2$ represents the uncertainty
of making a decision.  For the case when $\sigma^2 \rightarrow 0$ (no
uncertainty) we require the \texttt{nif} statement to behave as a
traditional \texttt{if} statement.

The evaluation of the \texttt{nif()} statement is explained on hand of
the following example, where \texttt{x}, \texttt{a} $\in \mathcal{R}$
and $\sigma^2$ $\in \mathcal{R^+}$.

\vspace{2mm}
\begin{small}
  \begin{lstlisting}[style=neural,mathescape]
nif( x >= a, $\sigma^2$) S1 else S2
\end{lstlisting}
\end{small}
\vspace{-1mm}

The evaluation is done in two steps: (i)~Find an $\mathcal{R}$
interval $I$ representing the confidence of making the decision;
(ii)~Check if a sample from the GD $\mathcal{N}(0,\sigma^2)$ belongs to $I$.

\begin{figure}[htbp]
\begin{center}
\includegraphics[width=0.95\linewidth]{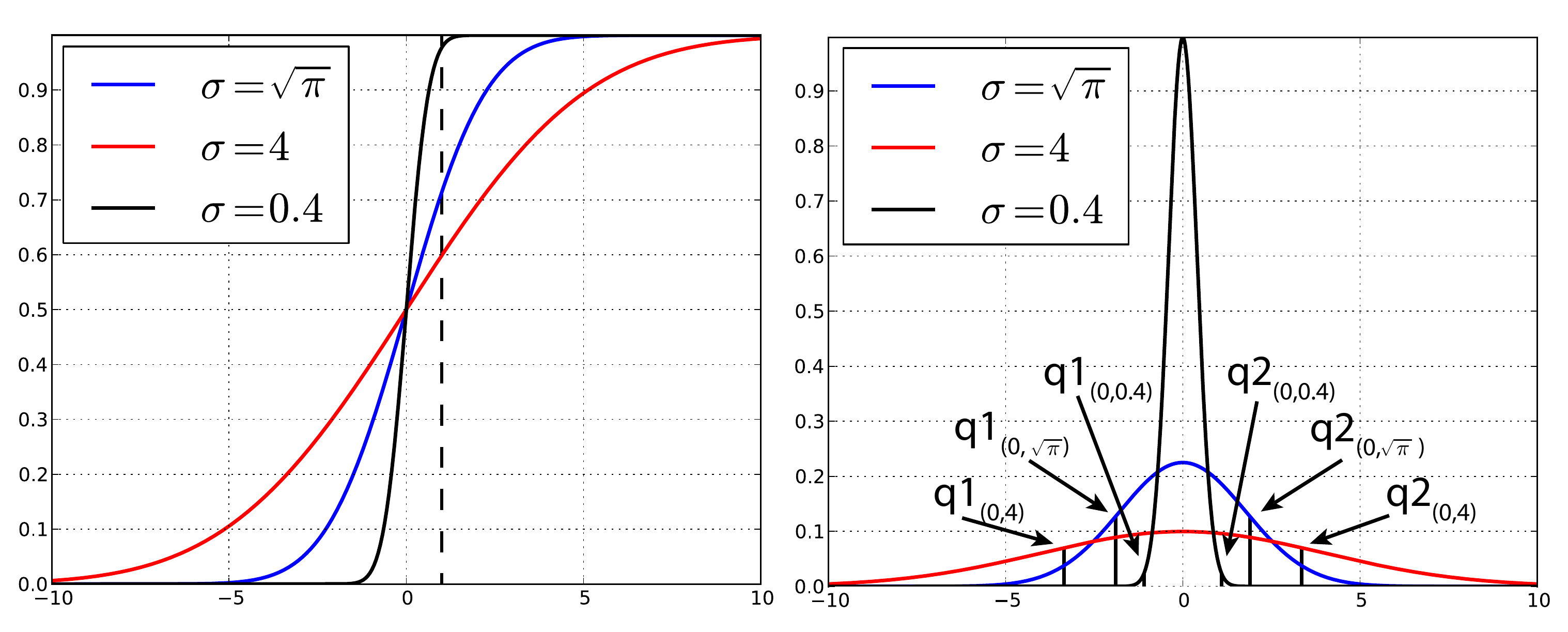}
\end{center}
\vspace*{-5mm}
\caption{CDFs, PDFs and the quantiles for $x = 1$ }
\label{fig:nwhile_sampling}
\end{figure}

Since the input RV has a GD, and a GD is used to evaluate the
condition, the result is a product of two GDs, which is also a GD
scaled by some constant factor $k$.  To find $I$ in (i), we estimate
the difference \texttt{diff(x,a)} between \texttt{x} and \texttt{a}.
For the general case \texttt{nif(x~$\#$~a,$\sigma^2$)}, with arbitrary
$\#$, the difference \texttt{diff(x,a)} is defined as below, where
$\epsilon$ represents the smallest real number on a computer.
\begin{equation*}
  \texttt{diff(x,a)} =
  \begin{cases} 
	\texttt{x - a}-\epsilon & \text{if}\; \#\; \text{is}\; >,\\
	\texttt{x - a}          & \text{if}\; \#\; \text{is}\; \geq,\\
	\texttt{a - x}-\epsilon & \text{if}\; \#\; \text{is}\;  <,\\
	\texttt{a - x}		  & \text{if}\; \#\; \text{is}\;  \leq.
  \end{cases} 
\end{equation*}
Informally, our confidence is characterized by the difference: The
larger \texttt{diff(x,a)} is, the larger is the probability of
executing \texttt{S1}.  The probability to execute \texttt{S1} is
given by $\texttt{cdf}_{0,\sigma^2}(\texttt{diff(x,a)})$ and
is used to obtain the interval $[q_1; q_2]$ by calculating two
symmetric quantiles $q_1$ and $q_2$ such that:
\begin{equation}
  \int_{q1}^{q2} \texttt{pdf}_{0,\sigma^2}(\texttt{x}) dx =
  \texttt{cdf}_{0, \sigma^2}(\texttt{diff(x,a)}).
  \label{eq:quantiles}
\end{equation}
In the second step a random sample from the distribution
$\mathcal{N}(0,\sigma^2)$ is checked to belong to the interval
$[q_1;q_2]$.  If it is within the interval, \texttt{S1} is executed,
otherwise \texttt{S2}.  At this point the probability value to execute
\texttt{S1} is influenced by the variance $\sigma^2$ (see
Figure~\ref{fig:nwhile_sampling}).
Hence, the dependence is twofold: \texttt{diff(x,a)} shows how
confident we are in making a decision, and $\sigma^2$
characterizes the uncertainty.

For the case $\sigma^2\rightarrow 0$ the \texttt{nif} statement is
equivalent to the \texttt{if} statement.  For infinitesimal $\sigma^2$
the PDF is expressed as a Dirac function $\delta(x)$, which has the
following properties: \\[2mm]
\hspace*{6mm}(i)\quad $\delta(x) = +\infty$ if $x = 0$ else $0$\\[1mm] 
\hspace*{6mm}(ii)\quad $\int_{-\infty}^\infty \delta(x)\,dx = 1$.\\[2mm]
The Dirac function essentially concentrates all the PD in a single
point $x\,{=}\,0$.  In this case the $\texttt{cdf}_{0,
  \sigma^2\rightarrow 0}(x)$ becomes a step function (see
Figure~\ref{fig:neuron}).  We consider two cases, as follows:
(i)~$\texttt{diff(x,a)}\,{\geq}\,0$ and
(ii)~$\texttt{diff(x,a)}\,{<}\,0$.  In the first case the probability
of executing \texttt{S1} is equal to $1$, hence the interval is
$(-\infty; +\infty)$ and includes every possible sample; for the
second case the probability of taking \texttt{S1} is $0$, hence the
interval is empty and cannot contain any sample.  An \texttt{if}
statement is an \texttt{nif} statement without uncertainty.

Let us illustrate the concept on a concrete example. Suppose that in
the current execution $\texttt{x} = 0$ and $\texttt{a}=1$.
Figure~\ref{fig:nwhile_sampling} illustrates how decisions are made if
$\sigma^2$ is: $0.4^2, \pi, 4^2$.
Since \texttt{diff(x,a) = 1}, the probability of executing \texttt{S1}
is defined by $\texttt{cdf}_{0,\sigma^2}(1)$ and for the above cases
is equal to 0.994, 0.714 and 0.599 respectively. The intervals for the
above cases are as follows: [-1.095;1.095], [-1.890;\,1.890] and
[-3.357;\,3.357].  In the second step we sample from the distributions
with the corresponding $\sigma^2$ ($\mathcal{N}(0,0.4^2)$,
$\mathcal{N}(0,\pi)$ and $\mathcal{N}(0,4^2)$) and check whether the
value lies within the intervals.\vspace*{1mm}

So far we were concerned with single samples $x \sim \mathcal{N}({\mu,
  \sigma^2})$.  Figure~\ref{fig:nif_cut} illustrates what happens to
the distributions: The differences of passing a GD RV \texttt{x $\sim$
  $\mathcal{N}(0,0.1)$ } through the statements \texttt{if(x >= 0.15)}
and \texttt{nif(x >= 0.15, 0.1)}.  Using our approach the GD is not
cut in undesirable ways, and it maintains its GD form after passing
the \texttt{nif} statement.\vspace*{1mm}

We can introduce now the confidence-uncertainty trade-off to loops.
The \emph{neural while} statement, or \texttt{nwhile} for short, is an
extension of a traditional \texttt{while} statement which incorporates
uncertainty. The statement
$\texttt{nwhile}(\;\texttt{x}~\#~\texttt{a}, \sigma^2) \{ P_1 \}$
takes an inequality relation and variance $\sigma^2$ and executes the
program $P_1$ according to the following rule: (1)~Compute
$\texttt{diff}(\texttt{x}~\#~\texttt{a})$ and obtain quantiles $q_1$
and $q_2$ according to the Equation~\ref{eq:quantiles}; (2)~Check if a
random sample $x \sim \mathcal{N}(0,\sigma^2)$ is within the interval
$[q_1; q_2]$; (3)~If the sample belongs to the interval, execute $P_1$
and go to step (1), else exit.

Since the \texttt{nif} and \texttt{nwhile} statements subsume the
behavior of traditional \texttt{if} and \texttt{while} statements (the
case $\sigma^2\rightarrow0$), we use them to \emph{define an
  imperative language with probabilistic control structures}. Binary
operators $bop$ (such as addition multiplication), unary operators
$uop$ (negation), and constants $c$ are used to form expressions
$E$. A program $P$ is a statement $S$ or combination of statements.\vspace*{1mm}
\begin{equation*}
  \begin{split}
	E ::=&\quad \texttt{x}_i\; |\; c\; | \; bop( E_1, E_2) \; | \; uop(E_1)\\
    S ::=&\quad \texttt{skip} \; | \; \texttt{x}_i := E \; | \; S_1; S_2
	\;| \; \texttt{nif} (\,\texttt{x}_i\, \# \, c , \sigma^2\,)\; S_1\; \texttt{else}\; S_2 \; | \\
	&\quad \texttt{nwhile}(\,\texttt{x}_i\, \# \, c,\,\sigma^2)\{\; S_1\; \}
  \end{split}
\end{equation*}
\vspace*{1mm}In order to define the denotational semantics for the \texttt{nif} and
the \texttt{nwhile} statements, we use
$\texttt{check}(\texttt{x}_i,c,\sigma^2, \#)$, a function which:
(1)~Computes the difference $\texttt{diff}(\texttt{x}_i, \texttt{c})$,
(2)~Finds quantiles $q_1$ and $q_2$ (Equation~\ref{eq:quantiles}), and
(3)~Checks if a sample $x~\sim~\mathcal{N}(0, \sigma^2)$ belongs to
the interval $[q_1;q_2]$.  If it does, it returns value $1$, otherwise
it returns value $0$.  The denotational semantics of neural programs
is then defined as follows:\vspace*{1mm}
\begin{equation*}
  \begin{split}
	&\llbracket \texttt{skip} \rrbracket(\texttt{x}) = \;\texttt{x} \\[4pt]
	&\llbracket \texttt{x}_i := E\;\rrbracket(\texttt{x}) = \;
	\texttt{x}[\llbracket E
	\rrbracket(\texttt{x})\mapsto\texttt{x}_i]\\[4pt]
	&\llbracket\; S_1; S_2\;\rrbracket(\texttt{x}) = \; \llbracket S_2
	\rrbracket (\llbracket S_1 \rrbracket (\texttt{x})) \\[4pt]
	&\llbracket \texttt{nif}(\,\texttt{x}_i\, \# \, c, \sigma^2)\;
	S_1\; \texttt{else}\; S_2\rrbracket(\texttt{x}) = \\
		&\quad\llbracket
		\texttt{check}(\texttt{x}_i,a,\sigma^2,\#)\rrbracket(\texttt{x})
		\llbracket S_1 \rrbracket(\texttt{x}) \quad{+} \\
		&\quad\llbracket
		\neg \texttt{check}(\texttt{x}_i,a,\sigma^2,\#)\rrbracket(\texttt{x})
		\llbracket S_2 \rrbracket(\texttt{x})\\[4pt]
 &\llbracket \texttt{nwhile}(\,\texttt{x}_i\, \# \, c,\,\sigma^2)\{\; S_1\; \}\rrbracket(\texttt{x}) = \\
	&\quad\texttt{x}\llbracket \neg \texttt{check}(\texttt{x}_i, a,
	\sigma^2, \#\,)\rrbracket(\texttt{x}) \quad{+}\\ 
	&\quad\llbracket\texttt{check}(\texttt{x}_i, a, \sigma^2,
	\#\,)\rrbracket(x) \llbracket \texttt{nwhile}(\,\texttt{x}_i\, \#
	\, c,\,\sigma^2)\{\; S_1\; \}\rrbracket(\llbracket S_1 \rrbracket
	\texttt{x})
  \end{split}
\end{equation*}

\vspace*{-3mm}We are now ready to write the control-program skeleton for the
parallel parking task of our Pioneer rover, as a sequence of
\texttt{nwhile} statements, as shown in Listing~\ref{lst:parking}.
Each \texttt{nwhile} corresponds to executing one motion primitive of
the informal description in Section~\ref{sec:introduction}. The
functions \texttt{moving()} and \texttt{getPose()} are output and
input statements,
which for simplicity, were omitted from the denotational semantics.

\begin{small}
  \begin{lstlisting}[caption={Parallel parking program
      skeleton\label{lst:parking}},style=neural,mathescape]
nwhile(currentDistance < targetLocation1, sigma1){
	moving(); 
	currentDistance = getPose();
	}
updateTargetLocations();
nwhile(currentAngle < targetLocation2,    sigma2){
	turning(); 
	currentAngle = getAngle();
	}
updateTargetLocations();
nwhile(currentDistance < targetLocation3, sigma3){
	moving(); 
	currentDistance = getPose();
	}
updateTargetLocations();
nwhile(currentAngle < targetLocation4,    sigma4){
	turning();
	currentAngle = getAngle();
	}
updateTargetLocations();
nwhile(currentDistance < targetLocation5, sigma5){
	moving();
	currentDistance = getPose();
	}
\end{lstlisting}
\end{small}
\vspace{-2mm}

The versatility of this approach is that the program skeleton is
written only once and comprises infinite number of controllers. The
question we need to answer next is:\\[2mm] 
\quad\emph{What are the distances and
  turning angles for each action and how uncertain are we about each
  of them?}\\[2mm]
To find the unknown parameters from Listing~\ref{lst:parking}, namely
the target locations \texttt{targetLocation}s and variances
\texttt{sigma}s, we use the learning procedure described in
Section~\ref{sec:learning}.

\vspace*{2mm}
\section{Bayesian-Network Learning}
\label{sec:learning}

Parking can be seen as a sequence of moves and turns, where each
action depends on the previous one. For example, the turning angle
typically depends on the previously driven distance.  Due to sensor
noise and imprecision, inertia and friction forces, and also many
possible ways to perform a parking task starting from one initial
location, we assume that the dependence between actions is
probabilistic, and in particular, the RVs are distributed according to
Gaussian distributions (GD).  We represent the dependencies between
actions as the GBN in Figure~\ref{fig:GBN}, where $l_i$ or $\alpha_j$
denotes a distance or a turning angle of the corresponding action and
$b_{ij}$ is a conditional dependence between consecutive actions.\vspace*{1mm}

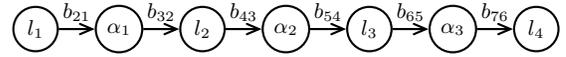
\begin{figure}[htbp]
\begin{center}
\begin{tikzpicture}[thick,scale=0.7, every node/.style={scale=0.9}]
\centering
\node [circle, draw] (loc){$l_1$};
\node [circle, draw] (loc1) [right=0.48cm of loc]{ $\alpha_1$};
\node [circle, draw] (loc2) [right=0.48cm of loc1]{ $l_2$};
\node [circle, draw] (loc3) [right=0.48cm of loc2]{ $\alpha_2$};
\node [circle, draw] (loc4) [right=0.48cm of loc3]{ $l_3$};
\node [circle, draw] (loc5) [right=0.48cm of loc4]{ $\alpha_3$};
\node [circle, draw] (loc6) [right=0.48cm of loc5]{ $l_4$};

\draw [->, -angle 45] (loc)  to node [above] {$b_{21}$} (loc1);
\draw [->, -angle 45] (loc1) to node [above] {$b_{32}$} (loc2);
\draw [->, -angle 45] (loc2) to node [above] {$b_{43}$} (loc3);
\draw [->, -angle 45] (loc3) to node [above] {$b_{54}$} (loc4);
\draw [->, -angle 45] (loc4) to node [above] {$b_{65}$} (loc5);
\draw [->, -angle 45] (loc5) to node [above] {$b_{76}$} (loc6);
\label{fig:GBN}
\end{tikzpicture}
\end{center}
\vspace*{-3mm}
\caption{Gaussian Bayesian Network for parking}
\label{fig:GBN}
\end{figure}

\begin{figure}[htbp]
\begin{center}
\includegraphics[width=0.9\linewidth]{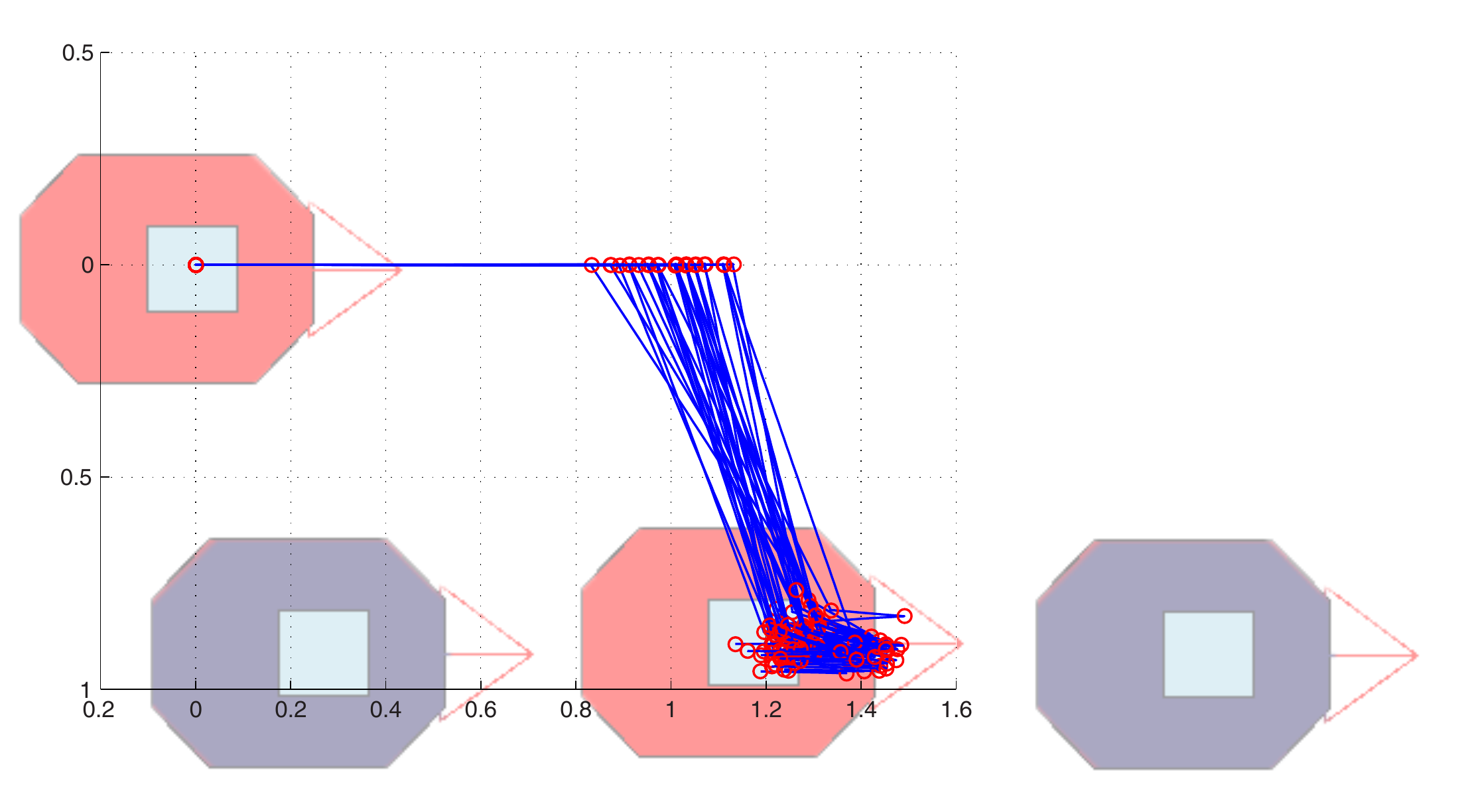}
\end{center}
\vspace*{-4mm}
\caption{Example trajectories for the parking task}
\label{fig:trajectories}
\end{figure}

In order to learn the conditional probability distributions of the GBN
in Figure~\ref{fig:GBN}, and to fill in the \texttt{targetLocation}s
and the \texttt{sigma}s in Listing~\ref{lst:parking}, we record
trajectories of the successful parkings done by a human expert.
Figure~\ref{fig:trajectories} shows example trajectories used during
the learning phase.\vspace*{1mm}

We than use the fact that any GBN can be converted to an
MGD~\cite{Neapolitan:2003} in our learning routine.
Learning the parameters of the GBN can be divided into three steps:
\begin{enumerate}
\vspace*{-2mm}\item Convert the GBN to the corresponding MGD,
\vspace*{-2mm}\item Update the precision matrix $\mathbf{T}$ of the MGD,
\vspace*{-2mm}\item Extract $\sigma^2$s and conditional dependences from $\mathbf{T}$.
\end{enumerate}
\vspace*{-2mm}{\em 1. Conversion step.}  To construct MGD we need to
obtain the mean vector $\mu$ and the precision matrix
$\mathbf{T}$. The mean vector $\mu$ comprises the means of all the
variables from the GBN.  To find the symbolic form of the precision
matrix, we use the recursive notation in \cite{HeckermanG95}, where
the value of the coefficients $b_i$, will be learned in the update
step below.\vspace*{-2mm}

\begin{equation}
  \mathbf{T}_{i+1} =\left( \begin{matrix}
	\textbf{T}_{i} +
	\frac{\mathbf{b}_{i+1}\mathbf{b}_{i+1}^T}{\sigma_{i+1}^2} &
	-\frac{\mathbf{b}_{i+1}}{\sigma_{i+1}^2} \\
	-\frac{\mathbf{b}_{i+1}^T}{\sigma_{i+1}^2} &
	\frac{1}{\sigma_{i+1}^2} \end{matrix}  \right)
  \label{eq:T}
\end{equation}
\vspace*{-2mm}

In order to apply Equation~\ref{eq:T} we define an ordering starting
with the initial node $l_1$. Its precision matrix is equal to:

\vspace*{-3mm}
\begin{equation*}
  T_{1} = \frac{1}{\sigma^2_1}.
\end{equation*}
\vspace*{-3mm}

The vector $\mathbf{b}_{i}$ comprises dependence coefficients for node
$i$ on all its immediate parents it in the ordering. For example, the
dependence vector for node $\alpha_2$ in the Figure~\ref{fig:GBN}
equals to:

\vspace*{-3mm}
\begin{equation*}
  b_{4} = \left(
  		\begin{matrix}
		0 \\
		0 \\
		b_{43} \\
		\end{matrix}
	  \right)
\end{equation*}
\vspace*{-3mm}

After applying the Equation~\ref{eq:T} to each node in the GBN, we
obtain the precision matrix $\mathbf{T}_7$, shown in
Figure~\ref{eq:precision}.  Since each action in the parking task
depends only on the previous one (for example, in Figure~\ref{fig:GBN}
the turning angle depends on the previously driven distance only), we
can generalize the precision matrix for the arbitrary number of
moves. For a GBN with $k$ moves, all non-zero elements of the
precision matrix $T \in \mathcal{R}^{k;k}$ can be found according to
the Equation~\ref{eq:T_gen}, where $\mathbf{T}(r,c)$ is a $c$-th
element in a $r$-th row of the precision matrix with indices started
from one.

\vspace*{-3mm}
\begin{equation}
  \begin{split}
  \mathbf{T}(i, i-1) =-\frac{b_{i(i-1)}}{\sigma^2_{i}},\\
  \mathbf{T}(i, i) =\frac{1}{\sigma^2_{i}} + \frac{b^2_{(i+1)i}}{\sigma^2_{i+1}},\\
  \mathbf{T}(i, i+1) =-\frac{b_{(i+1)i}}{\sigma^2_{i+1}},
  \end{split}
  \label{eq:T_gen}
\end{equation}
\vspace*{-3mm}

\emph{2. Update step.}  Once we derived the symbolic form of the
precision matrix ($\mathbf{T}_7$ in our example), we use the training
set, in order to learn the actual values of its parameters, as
described in the algorithm from \cite{Neapolitan:2003}.  Each training
example $\mathbf{x}^{(i)}$ corresponds to a vector of lengths and
turning angles for a successful parking task. The total number of
examples in the training set is $M$.
The procedure allows us to learn iteratively and adjust the prior
belief by updating the values of the mean $\mu$ and covariance matrix
$\beta$ of the prior, where $v$ is a size of a training set for the
prior belief, and $\alpha = v-1$.

\vspace*{-0.45cm}
\begin{equation} 
  \beta = \frac{v(\alpha - n + 1)}{v + 1} \mathbf{T}^{-1},
\end{equation}
\vspace*{-0.45cm}

The updated mean value $\mu^*$ incorporates prior value of the mean
$\mu$ and the mean value of the new training examples $\mathbf{x}$.

\vspace*{-0.45cm}
\begin{equation} 
  \begin{split}
  \overline{\mathbf{x}} =& \frac{\sum_{i =1}^M \mathbf{x}^{(i)}}{M} \\
  \mu^* =& \frac{v\mu + M\overline{\mathbf{x}}}{v+ M}
  \end{split}
\label{eq:x_and_mu}
\end{equation}
\vspace*{-0.45cm}

The size of the training set $v^*$ is updated to its new value:

\vspace*{-0.6cm}
\begin{equation} 
  v^* = v + M
\end{equation}
\vspace*{-0.6cm}

The updated covariance matrix $\beta^*$ combines the prior matrix
$\beta$ with the covariance matrix of the training set $\mathbf{s}$:

\vspace*{-0.5cm}
\begin{equation} 
  \begin{split}
  \mathbf{s} =& \sum_{i =1}^M \left(x^{(i)} - \overline{\mathbf{x}}\right)\left(x^{(i)} - \overline{\mathbf{x}}\right)^T \\
  \beta^* =& \beta + s + \frac{rm}{v+M}\left(x^{(i)} - \overline{\mathbf{x}}\right)\left(x^{(i)} - \overline{\mathbf{x}}\right)^T
  \end{split}
\end{equation}
\vspace*{-0.3cm}

Finally, the new value of the matrix $\beta$ is used to calculate
the covariance matrix $(\mathbf{T}^*)^{-1}$, where $\alpha^*=\alpha+M$.

\vspace*{-0.4cm}
\begin{equation} 
  {(\mathbf{T}^*)}^{-1} = \frac{v^* + 1}{v^*(\alpha^* - n + 1) } \beta^*
  \label{eq:T_new}
\end{equation}
\vspace*{-0.2cm}
\begin{figure*}[!t]
\begin{equation}
  \textbf{T}_7 =\left( 
	    \begin{matrix}
	      \frac{1}{\sigma^2_1} + \frac{b_{21}^2}{\sigma^2_2} & -\frac{b_{21}}{\sigma_2^2} & 0 & 0 & 0 & 0 & 0\\
	      -\frac{b_{21}}{\sigma_2^2} & \frac{1}{\sigma^2_2} + \frac{b_{32}^2}{\sigma_3^2} & -\frac{b_{32}}{\sigma_3^2} 
	      & 0 & 0 & 0 & 0\\
	      0 & -\frac{b_{32}}{\sigma_3^2} & \frac{1}{\sigma_3^2} + \frac{b_{43}^2}{\sigma_4^2} & -\frac{b_{43}}{\sigma_4^2}
	      & 0  & 0 & 0\\
	      0 & 0 & -\frac{b_{43}}{\sigma_4^2} & \frac{1}{\sigma_4^2} +\frac{b_{54}^2}{\sigma_5^2} & 
	      -\frac{b_{54}}{\sigma_5^2} & 0 & 0\\
		  0 & 0 & 0 &-\frac{b_{54}}{\sigma_5^2} & \frac{1}{\sigma_5^2} + \frac{b_{65}^2}{\sigma_6^2} & -\frac{b_{65}}{\sigma_6^2} & 0\\
	      0 & 0 & 0 & 0 & -\frac{b_{65}}{\sigma_6^2} & \frac{1}{\sigma_6^2} + \frac{b_{76}^2}{\sigma_7^2} & -\frac{b_{76}}{\sigma_7^2}\\
	      0 & 0 & 0 & 0 & 0 & -\frac{b_{76}}{\sigma_7^2} & \frac{1}{\sigma_7^2} \\
	    \end{matrix}  
	    \right)
\label{eq:precision}
\end{equation}
\vspace*{-2mm}
\end{figure*}

\begin{figure}[htbp]
\begin{center}
\includegraphics[width=0.9\linewidth]{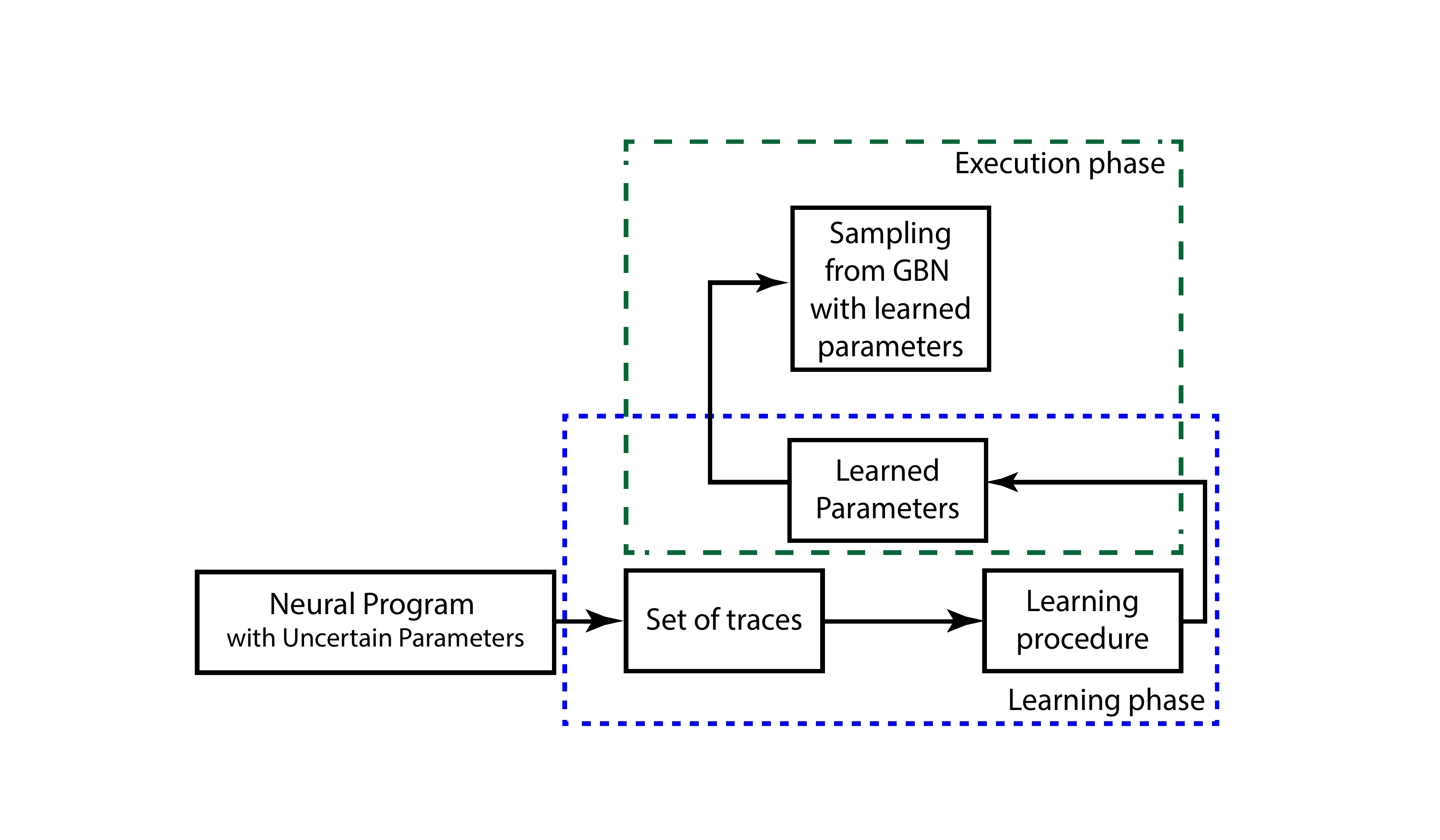}
\end{center}
\vspace*{-3mm}
\caption{Learning Parameters in a Neural Program}
\label{fig:2phaseProgram}
\vspace*{2mm}
\end{figure}

\emph{3. Extraction step.} The new parameters of the GBN can now be
retrieved from the updated mean vector $\mu^*$ and from
$(\mathbf{T}^*)^{-1}$.  If new traces are available at hand, one can
update the distributions by recomputing $\mu^*$ and
$(\mathbf{T}^*)^{-1}$ using
Equations~\ref{eq:x_and_mu}-\ref{eq:T_new}.  We depict the whole
process in Figure~\ref{fig:2phaseProgram}: Unknown parameters from
the program skeleton are learned from successful traces and these
dependencies are used during the execution phase to sample the
commands.\vspace*{2mm}

\vspace*{4mm}\section{Experimental results}
\label{sec:experiments}

\begin{figure}[htbp]
\begin{center}
\includegraphics[width=0.8\linewidth]{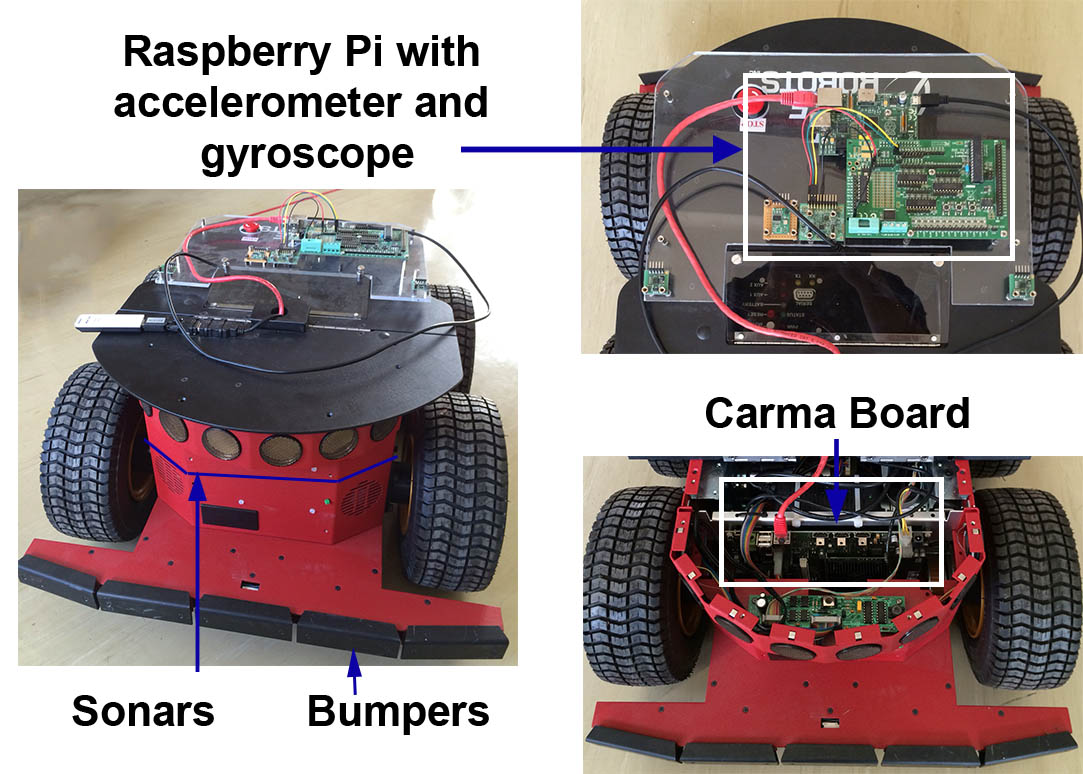}
\end{center}
  \vspace*{-2mm}
\caption{Experimental platform: Pioneer Rover}
\label{fig:robot}
\end{figure}

We performed our experiments on a \texttt{Pioneer P3AT-SH} mobile
rover from Adept MobileRobots\cite{Adept} (see
Figure~\ref{fig:robot}).  The rover uses the \texttt{Carma Devkit}
from SECO \cite{CarmaPoster} as a main computational unit. The
comprised Tegra 3 ARM CPU runs the Robot Operating System (ROS) on top
of Ubuntu 12.04.\vspace*{1mm}

\vspace*{2mm}
\subsection{Structure of the Parking System}
\label{subsec:structure}
\begin{figure}[htbp]
\begin{center}
\includegraphics[width=\linewidth]{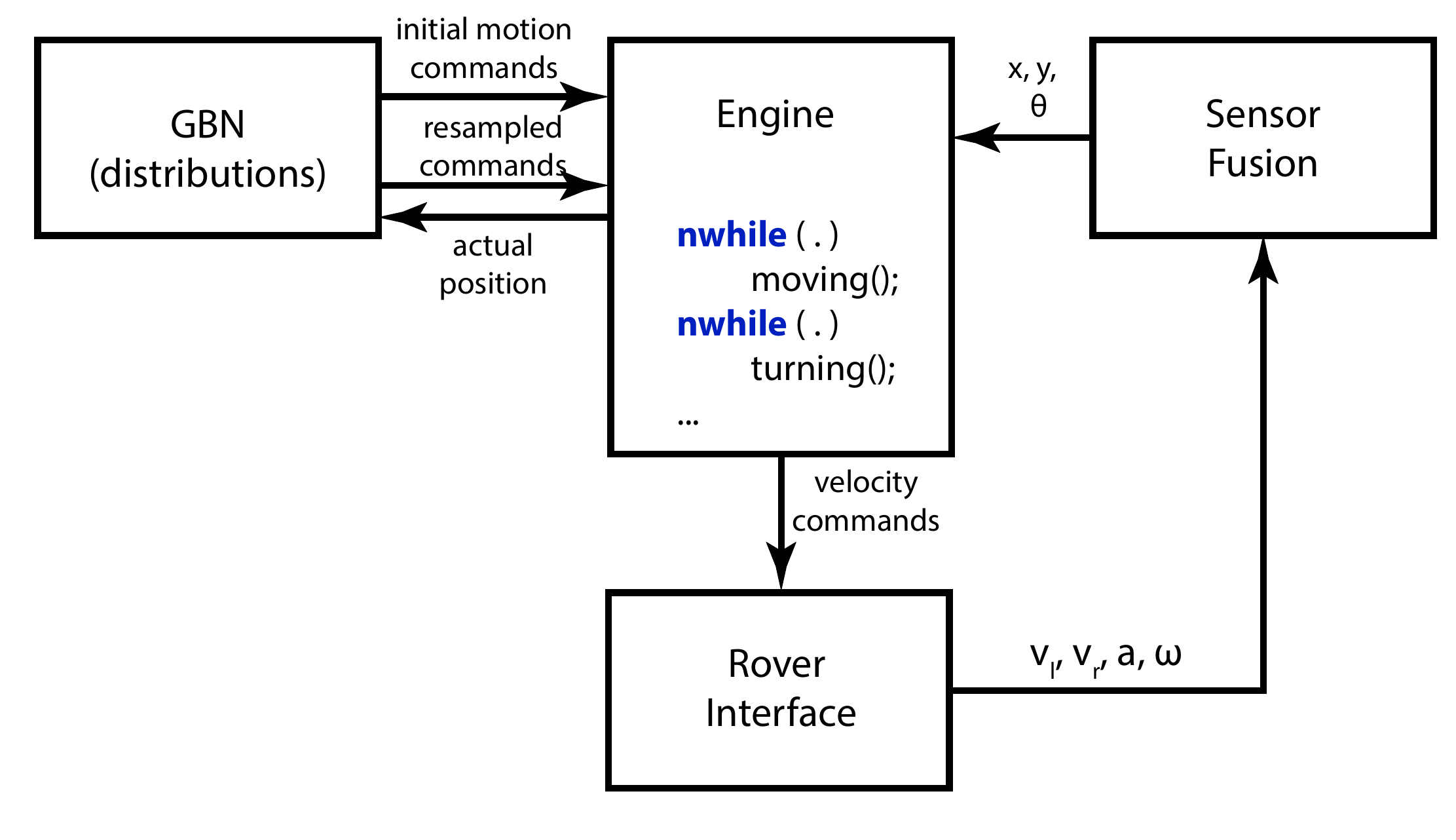}
\end{center}
  \vspace*{-6mm}
\caption{Parking system architecture}
\label{fig:parking_sys}
\end{figure}

The parking system can be separated into several building blocks (see
Figure \ref{fig:parking_sys}). The block \emph{Rover Interface} senses
and controls the rover, that is, it establishes an interface to the
hardware. The block \emph{Sensor Fusion} takes the sensor values from
the \emph{Rover Interface} block, and provides the estimated pose of
the rover to the high-level controller \emph{Engine}. The
\emph{Engine} uses the \emph{GBN} block to update the motion commands
based on the estimated pose. Furthermore, the \emph{Engine} maps the
(higher level) motion commands to velocity commands needed by the
\emph{Rover Interface} to control the rover.

\vspace*{2mm}
\subsubsection{The Gaussian Bayesian Network Block}

The goal of the GBN block in Figure~\ref{fig:parking_sys}, is to
generate motion commands for the Engine to execute. A motion command
corresponds to a driving distance or a turning angle.

During the learning phase, the distributions of the random variables
(RVs) in the Gaussian Bayesian network (GBN) in Figure \ref{fig:GBN}
are collected in a CSV file of following format:\vspace*{1mm}
\lstset{basicstyle=\scriptsize}
\begin{lstlisting}[frame=single] 
motionType,motionDirection,mean,variance,
dependenceCoefficient
\end{lstlisting}
\vspace*{-3mm}
Parsing the CSV file initializes the GBN that will be used for
sampling the motion commands. Before starting the run we obtain the
initial command vector from the distributions learned.  The
distribution of the first move $l_1$ is independent from any other
move and has the form $\mathcal{N}(\mu_1, \sigma_1^2)$. Starting from
the second move $\alpha_1$, each motion depends on the previous one:
For motion number $n$, the distribution has the form
$\mathcal{N}(\mu_n\,{-}\,b_{n,n-1}\,{*}\,~x^{\text{\tiny{sampled}}}_{n-1},
\sigma_n^2)$.  The initial command vector is obtained by sampling from
$l_1$, and each subsequent command vector is obtained by taking into
account the previous sample, when sampling from its own distribution.

As the rover and its environment are uncertain (e.g., sensors are
disturbed by noise) we use the pose provided by the sensor fusion unit
to update the motion commands. Hence the motion commands are
constantly adapted to take into account the actual driven distance
(which could be different from the planned one due to the
aforementioned uncertainty of the CPS). This allows us to incorporate
the results of the sensor fusion algorithm in the updated commands.

\vspace*{2mm}
\subsubsection{The Engine Block}
\label{subsec:engine}

During the run we execute a motion command according to the
semantics of the \texttt{nwhile} loop. 
In particular, the estimated pose is passed from the Sensor Fusion
block to the Engine and compared with the target location, specified
as a point on a 2-D plane.  Since the rover is affected by noise its
path can deviate and never come to the target location. To be able to
detect and overcome this problem we estimate the scalar product of two
vectors: The first one is the initial target location, and the second
one is the current target location.  This product is monotonely
decreasing and becomes negative after passing the goal even on a
deviating path.  In an \texttt{nwhile} statement we monitor the
distance (or angle) and detect if we should process the next command.

To obtain the current state of the rover (its pose), and send velocity
commands, we start two separate threads: (i)~Receive the pose
and (ii)~Send the velocity command.  The motion command
(containing desired driving distance or turning angle) is converted to
a suitable velocity or steering command respectively, for the
\emph{Rover Interface}.  After each executed command, we resample
the pose in order to take into account actual driving distance in
subsequent moves.

\vspace*{2mm}
\subsubsection{The Rover Interface Block}
The block \emph{Rover Interface} implements the drivers for sensors
and actuators. The \emph{wheel velocities} are measured by encoders,
already supplied within the Pioneer rover. A built-in microcontroller
reads the encoders and sends their value to the \texttt{Carma
  Devkit}. Additionally the rover is equipped with an \emph{inertial
  measurement unit (IMU)} including an accelerometer, measuring the
linear acceleration, and a gyroscope, measuring the angular velocity
of the rover. The \texttt{Raspberry Pi} mounted on top of the rover
samples the accelerometer and gyroscope, and forwards the raw IMU
measurements to the \texttt{Carma Devkit}. The rover is controlled
according to the incoming velocity commands containing desired linear
velocity into forward direction (x-translation) and the desired
angular velocity (z-rotation). The desired translation and rotation is
converted to the individual wheel velocities, which are sent to and
applied by the built-in microcontroller.

\vspace*{2mm}
\subsubsection{The Sensor Fusion Block}
Parking is often performed by applying predefined rules
\cite{LiYing10} or following a specific trajectory \cite{LiYing10}. So
typically an autonomously parking car stops at a specific position
beside a parking space and then turns and moves for a fixed time,
angle or distance. The car has to stop, move and turn \emph{exactly}
as designated to park successfully. The car has to be aware of its
current pose, that is, position and heading, otherwise parking will
most likely fail (whatsoever controller is used). However, the current
pose is observed by sensors, which suffer from
uncertainty. Measurements are distorted by noise, e.g., caused by
fluctuations of the elements of the electrical circuit of the
sensors. The environment may be unpredictable, e.g., the car may slip
over water or ice when parking. To overcome such problems sensor
fusion techniques are applied, i.e., several sensors are combined to
estimate a more accurate state. A common method is state estimation
(also called \emph{filtering}) \cite{Mit07,Thr06}.

In this application, an \emph{unscented Kalman filter (UKF)}
\cite{Wan00} is used. This filter combines the measurements listed in
Table \ref{tab:sensors} with a suitable model describing the relations
from the measured variables to the pose of the car.\vspace*{2mm}
\begin{table}[h!]
  \renewcommand{\arraystretch}{1.05}
  \centering
  \begin{tabular}{|c|p{0.22\textwidth}|c|}
    \hline
    & \textsc{Sensor} & \textsc{Variance}  \\ 
    \hline \hline
    $v_l$ & left wheel's velocity  & 0.002 $m/s$ \\
    $v_r$ & right wheel's velocity & 0.002 $m/s$ \\
    $a$ & linear acceleration & 0.25 $m/s^2$ \\
    $\omega$ & angular velocity & 0.00017 $rad/s$ \\
    \hline
  \end{tabular}
\vspace*{-1mm}
  \caption{Used sensors and its variances.}
  \label{tab:sensors}
\vspace*{-1mm}
\end{table}

The \emph{belief state} maintained by the UKF, e.g., the current
linear velocity, will be continuously updated: (i) By predicting
the state, and (ii) By updating the prediction with
measurements. For example, the linear velocity will be predicted by
the current belief of acceleration and the time elapsed since the
previous estimation. Next, the measurements from accelerometer and
wheel encoders are used to update the predicted velocity. Because the
wheel encoders are much more accurate than the acceleration sensor
(see variance in Table \ref{tab:sensors}), the measurements from the
wheel encoders will be trusted more (for simplicity one can think of
weighting and averaging the measurements, where the particular weight
corresponds to the reciprocal variance of the sensor). However, by
using more than one sensor, the unscented Kalman filter reduces the
error of estimated and actual velocity.

\vspace*{2mm}
\subsection{Integration into ROS}
\label{subsec:ROSintegration}

ROS~\cite{ROS2009} is a meta-operating system that provides common
functionality for robotic tasks including process communication,
package management, and hardware abstraction. A basic building block
of a system in ROS is a so-called \emph{node}, that performs
computation and exchanges information with other entities. Nodes
communicate with each other subscribing for or publishing messages to
specific \emph{topics}. So all the nodes subscribed to a particular
topic A, will receive messages from nodes publishing to this topic A.

\begin{figure}[htbp]
  \centering
  \includegraphics[width=\linewidth]{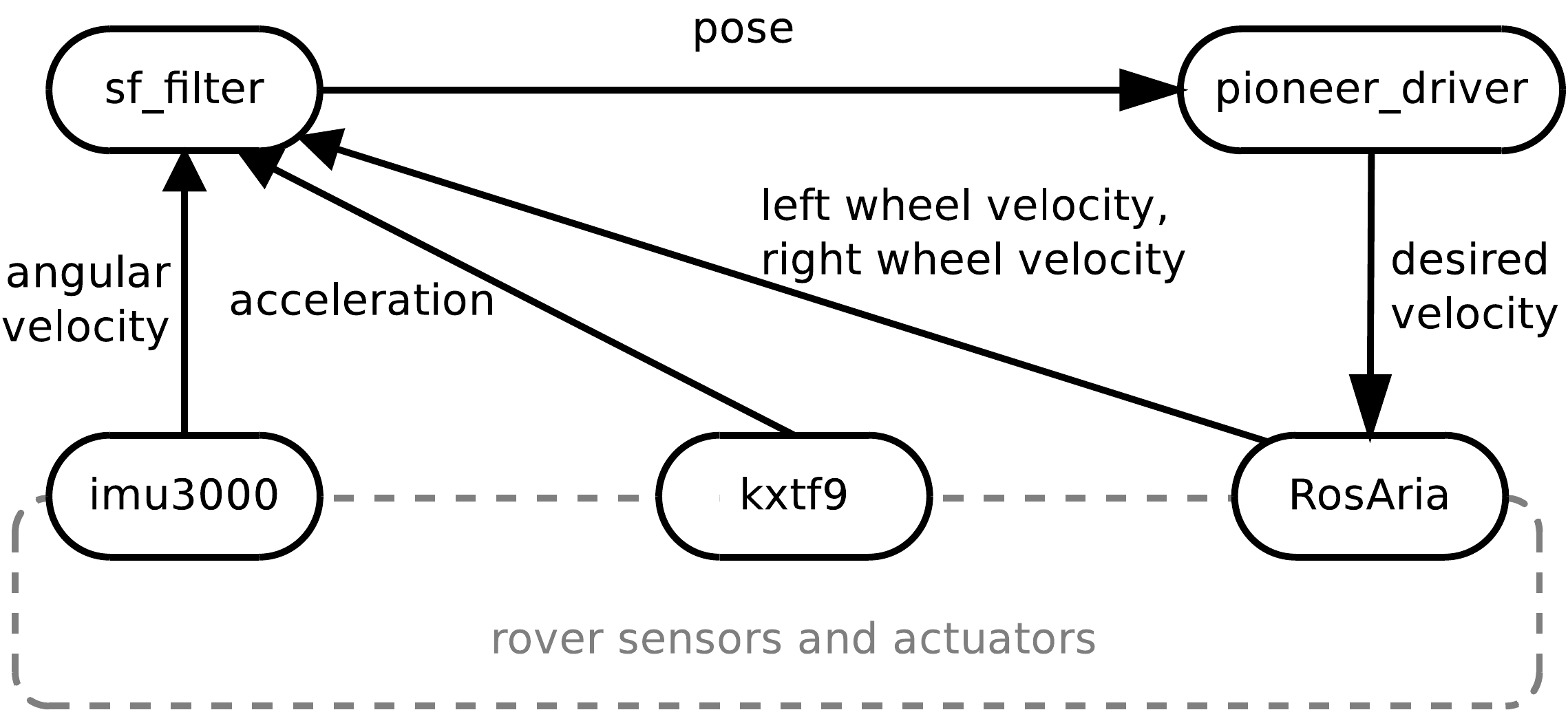}
  \vspace*{-4mm}
  \caption{Parking system in ROS.}
  \label{fig:parking_sys_ros}
\end{figure}

Since the application is implemented in ROS we use the utility
\texttt{roslaunch} to start the required ROS nodes (as shown in Figure
\ref{fig:parking_sys_ros}) corresponding to the blocks given in
Section \ref{subsec:structure}.
\begin{description}
  \vspace*{-4mm}\item[Rover Interface:] The node \texttt{RosAria} is
  used to control the velocity of the rover and provide the values of
  the wheel encoders for the sensor fusion node. The ROS nodes
  \texttt{imu3000} and \texttt{kxtf9} running on the \texttt{Raspberry
    Pi} provide data from acceleromenter and gyroscope.
  \vspace*{-2mm}\item[Sensor Fusion:] \texttt{sf\_filter} node reads
  sensor values, implements the sensor fusion algorithm and provides
  the estimated pose of the rover.  \vspace*{-2mm}\item[GBN and
  Engine:] \texttt{pioneer\_driver} is a node implementing resampling
  of commands based on the actual driven distance and constantly
  providing the required velocity commands to the \texttt{RosAria}
  node (see Figure \ref{fig:parking_sys_ros}).
\end{description}
%

\vspace*{2mm}
\subsection{Results}

After the learning phase, we obtain the parameters of the GBN that we
use in the program skeleton (Table.~\ref{tab:GBN_val}).  Since we
track the position using the data from the sensor fusion and each
movement has the experimentally learned uncertainty, we are resistive
to the perturbation of the actual driving distances and angles. If in
the current distance of the robot deviates from the planned one, the
commands, resampled from the GBN will try compensate the deviation
with the dependencies obtained from the learning phase.\vspace*{1mm}

\begin{table}[h!]
  \renewcommand{\arraystretch}{1.05}
  \centering
  \begin{tabular}{|c|c|c|c|c|c|c|c|}
    \hline
$-$	         	 &  	   & $b_{21}$ 	  & 0.7968 & $b_{32}$     & -0.2086  & $b_{43}$ & 0.5475 \\ \hline
$\sigma_1^2$	 & 0.0062  & $\sigma_2^2$ & 0.0032 & $\sigma_3^2$ & 0.0019   & $\sigma_4^2$ & 0.022  \\
\hline \hline
$b_{54}$	   	 & -0.0045 & $b_{65}$ 	  & 1.1920 & $b_{76}$     & -0.0968  &  & \\ \hline
$\sigma_5^2$	 & 0.0008  & $\sigma_6^2$ & 0.0178 & $\sigma_7^2$ & 0.0013   &  &   \\
    \hline
  \end{tabular}
  \caption{BN variances and coefficient dependences.}
  \label{tab:GBN_val}
\vspace*{4mm}
\end{table}


\vspace*{2mm}
\section{Related Work}
\label{sec:related}

Although probabilistic programs (PP), Gaussian Bayesian networks (GBN)
and neural networks were considered before, to the best of our
knowledge, the development of smooth control statements, related
within a GBN ontology is new. The ontology represents the knowledge of
the PP about both its environment and its own control logic.

Probabilistic programs are represented by different languages and
frameworks~\cite{Gilks94, RajamHenz14, Dekhtyar00}. The authors
in~\cite{RajamHenz14} differentiate probabilistic programs from
``traditional'' ones, by the ability to sample at random from the
distribution and condition the values of variables via
observation. Although they consider both discrete and continuous
probability distributions, and transformation of Bayesian Networks and
Discrete Time Markov Chains to probabilistic programs, they do not
mention probabilistic control structures linked in GBN.

In~\cite{Chaudhuri10}, the authors adapted the signal and image
processing technique called \emph{Gaussian smoothing (GS)}, for
program optimization. Using GS, a program could be approximated by a
smooth mathematical function, which is in fact a convolution of a
denotational semantics of a program with a Gaussian function. This
approximation is used to facilitate optimization for solving the
parameter synthesis problem.  In~\cite{Chaudhuri11} this idea was
developed further and soundness and robustness for smooth
interpretation of programs was defined.  In both papers the authors
do not consider any means for eliminating the re-normalization step
of the probability density function when a variable is passed through
a conditional branch in the current execution trace. Moreover, they
stop short of proposing new, smooth control statements.

Learning Bayesian Networks comprises different tasks and problem
formulations: $i)$~Learning the structure of the network,
$ii)$~Learning the conditional probabilities for the given structure
and $iii)$ Performing querying-inference for a given Bayesian
Network~\cite{Neapolitan:2003}. In~\cite{HeckermanG95} the authors
introduce a unified method for both discrete and continuous domains
to learn the parameters of Bayesian Network, using a combination of
prior knowledge and statistical data.

Various formulations of a mobile parking problem were extensively
studied for robots with different architectures~\cite{Manzan12,
  Jiang99, Khoshe05, Sciclu12, LiYing10}. For instance,
in~\cite{LiYing10} the authors use a custom spatial configuration of
the ultrasonic sensors and binaural method to perceive the environment
and park the robot using predefined rules.  In~\cite{Khoshe05} the
authors approximate the trajectory for the parking task with a
polynomial curve, that the robot could follow with the constraints
satisfied, and used fuzzy controller to minimize the difference
between specified trajectory and actual path.

In order to govern a physical process (e.g., parking a car), the
controller must be aware of the internal state of the process (e.g.,
the position of the car). Sensors measure the outputs of a process,
whereof the state can be estimated. However, the measurements are
distorted by noise and the environment may be unpredictable. State
estimators \cite{Mit07,Thr06,Aru02} and in particular Kalman filters
\cite{Thr06,Wan00} are commonly used methods to increase the
confidence of the state estimate evaluated out of raw sensor
measurements.


\vspace*{2mm}
\section{Conclusion}
\label{sec:conclusion}

In this paper we introduced \emph{deep neural programs (DNP)}, a new
formalism for writing robust and adaptive cyber-physical-system (CPS)
controllers. Key to this formalism is: (i) The use of smooth Probit
distributions in conditional and loop statements, instead of their
classic stepwise counterparts, and (2) The use of a Gaussian Bayesian
network, for capturing the dependencies among the Probit distributions
within the conditional and loop statements in the DPN.

We validated the usefulness of DPNs by developing, once and for all, a
parallel parking CPS-controller, which is able to adapt to unforeseen
environmental situations, such as a slippery ground, or noisy
actuators. No classic program has such ability: One would have to
encode all this unforeseen situations, which would lead to an
unintelligible code.

In future work, we plan to explore the advantages of DPNs in the
analysis, as well as, in the design (optimization) of CPS
controllers. The nice mathematical properties of DPNs make them an
ideal formalism for these tasks.

\vspace*{2mm}
\bibliographystyle{abbrv}
\bibliography{docear,grosu}  

\begin{thebibliography}{10}

\bibitem{Adept}
{Adept MobileRobots(2013). Pioneer 3-AT.}
\newblock
  \href{http://www.mobilerobots.com/ResearchRobots/P3AT.aspx}{http://www.mobilerobots.com/ResearchRobots/
  P3AT.aspx} \textit{(Accessed 24.08.2014)}.

\bibitem{neuralVideos}
{Parking Videos}.
\newblock
  \href{http://youtu.be/xNOj\_ARSEYs?list=PLP5Gx6r7gK2cxjKv0K2V5fBedovfo8\_3y}{http://youtu.be/xNOj\_
  ARSEYs?list=PLP5Gx6r7gK2cxjKv0K2V5fBedovfo8\_3y} \textit{(Accessed
  12.10.2014)}.

\bibitem{Aru02}
M.~Arulampalam, S.~Maskell, N.~Gordon, and T.~Clapp.
\newblock {A Tutorial on Particle Filters for Online Nonlinear/Non-Gaussian
  Bayesian Tracking}.
\newblock {\em IEEE Transactions on Signal Processing}, 50(2):174--188, 2002.

\bibitem{broyCPS}
M.~Broy and E.~Geisberger.
\newblock Cyber-physical systems, driving force for innovation in mobility,
  health, energy and production.
\newblock {\em Acatech: The National Academy Of Science and Engineering}, 2012.

\bibitem{LezamaSlides10}
S.~Chaudhuri and A.~Solar-Lezama.
\newblock {Smooth Interpretation: Presentation Slides}.
\newblock
  \href{http://people.csail.mit.edu/asolar/Talks/PLDI2010Final.pptx}{http://people.csail.mit.edu/asolar/Talks/
  PLDI2010Final.pptx} \textit{(Accessed 14.06.2014)}.

\bibitem{Chaudhuri10}
S.~Chaudhuri and A.~Solar-Lezama.
\newblock Smooth interpretation.
\newblock In {\em PLDI}, pages 279--291, 2010.

\bibitem{Chaudhuri11}
S.~Chaudhuri and A.~Solar-Lezama.
\newblock Smoothing a program soundly and robustly.
\newblock In {\em CAV}, pages 277--292, 2011.

\bibitem{Ciresan12}
D.~Ciresan, U.~Meier, and J.~Schmidhuber.
\newblock Multi-column deep neural networks for image classification.
\newblock In {\em Computer Vision and Pattern Recognition (CVPR), 2012 IEEE
  Conference on}, pages 3642--3649, June 2012.

\bibitem{Dekhtyar00}
A.~Dekhtyar and V.~Subrahmanian.
\newblock {Hybrid Probabilistic Programs }.
\newblock {\em The Journal of Logic Programming}, 43(3):187 -- 250, 2000.

\bibitem{Erhan10}
D.~Erhan, Y.~Bengio, A.~Courville, P.-A. Manzagol, P.~Vincent, and S.~Bengio.
\newblock Why does unsupervised pre-training help deep learning?
\newblock {\em J. Mach. Learn. Res.}, 11:625--660, Mar. 2010.

\bibitem{Fraenzle99}
M.~Fraenzle.
\newblock Analysis of hybrid systems: An ounce of realism can save an infinity
  of states.
\newblock In {\em Computer Science Logic}, volume 1683 of {\em Lecture Notes in
  Computer Science}, pages 126--139. Springer Berlin Heidelberg, 1999.

\bibitem{Gao14}
S.~Gao, S.~Kong, W.~Chen, and E.~M. Clarke.
\newblock Delta-complete analysis for bounded reachability of hybrid systems.
\newblock {\em CoRR}, abs/1404.7171, 2014.

\bibitem{Gilks94}
W.~R. Gilks, A.~Thomas, and D.~J. Spiegelhalter.
\newblock {A Language and Program for Complex Bayesian Modelling}.
\newblock {\em Journal of the Royal Statistical Society. Series D (The
  Statistician)}, 43(1):pp. 169--177, 1994.

\bibitem{RajamHenz14}
A.~D. Gordon, T.~A. Henzinger, A.~V. Nori, and S.~K. Rajamani.
\newblock {Probabilistic Programming}.
\newblock In {\em International Conference on Software Engineering (ICSE Future
  of Software Engineering)}. IEEE, May 2014.

\bibitem{grimmett_book}
G.~Grimmett and D.~Stirzaker.
\newblock {\em Probability and random processes}.
\newblock Oxford science publications. Clarendon Press, 1985.

\bibitem{HeckermanG95}
D.~Heckerman and D.~Geiger.
\newblock Learning bayesian networks: A unification for discrete and gaussian
  domains.
\newblock In {\em UAI}, pages 274--284, 1995.

\bibitem{Manzan12}
M.-A. Ibarra-Manzano, J.-H. De-Anda-Cuellar, C.-A. Perez-Ramirez, O.-I.
  Vera-Almanza, F.-J. Mendoza-Galindo, M.-A. Carbajal-Guillen, and D.-L.
  Almanza-Ojeda.
\newblock Intelligent algorithm for parallel self-parking assist of a mobile
  robot.
\newblock In {\em Electronics, Robotics and Automotive Mechanics Conference
  (CERMA), 2012 IEEE Ninth}, pages 37--41, Nov 2012.

\bibitem{Jiang99}
K.~Jiang and L.~Seneviratne.
\newblock A sensor guided autonomous parking system for nonholonomic mobile
  robots.
\newblock In {\em Robotics and Automation, 1999. Proceedings. 1999 IEEE
  International Conference on}, volume~1, pages 311--316 vol.1, 1999.

\bibitem{Khoshe05}
M.~Khoshnejad and K.~Demirli.
\newblock Autonomous parallel parking of a car-like mobile robot by a
  neuro-fuzzy behavior-based controller.
\newblock In {\em Fuzzy Information Processing Society, 2005. NAFIPS 2005.
  Annual Meeting of the North American}, pages 814--819, June 2005.

\bibitem{Koller09}
D.~Koller and N.~Friedman.
\newblock {\em Probabilistic Graphical Models: Principles and Techniques -
  Adaptive Computation and Machine Learning}.
\newblock The MIT Press, 2009.

\bibitem{CarmaPoster}
M.~A. Lee.
\newblock {CUDA on ARM: Tegra3 Based Low-Power GPU Compute Node}, 2013.
\newblock {Poster presented at GPU Technical Conference, 2013}.

\bibitem{LiYing10}
T.~Li, Y.-C. Yeh, J.-D. Wu, M.-Y. Hsiao, and C.-Y. Chen.
\newblock {Multifunctional Intelligent Autonomous Parking Controllers for
  Carlike Mobile Robots}.
\newblock {\em Industrial Electronics, IEEE Transactions on}, 57(5):1687--1700,
  May 2010.

\bibitem{Mit07}
H.~Mitchell.
\newblock {\em {Multi-Sensor Data Fusion - An Introduction}}.
\newblock Springer, Berlin, Heidelberg, New York, 2007.

\bibitem{Neapolitan:2003}
R.~E. Neapolitan.
\newblock {\em Learning Bayesian Networks}.
\newblock Prentice-Hall, Inc., Upper Saddle River, NJ, USA, 2003.

\bibitem{Parnas85}
D.~L. Parnas.
\newblock Software aspects of strategic defense systems.
\newblock {\em Commun. ACM}, 28(12):1326--1335, Dec. 1985.

\bibitem{ROS2009}
M.~Quigley, K.~Conley, B.~Gerkey, J.~Faust, T.~B. Foote, J.~Leibs, R.~Wheeler,
  and A.~Y. Ng.
\newblock {ROS}: an open-source robot operating system.
\newblock In {\em ICRA Workshop on Open Source Software}, 2009.

\bibitem{Ratschan06}
S.~Ratschan and Z.~She.
\newblock Constraints for continuous reachability in the verification of hybrid
  systems.
\newblock In J.~Calmet, T.~Ida, and D.~Wang, editors, {\em Artificial
  Intelligence and Symbolic Computation}, volume 4120 of {\em Lecture Notes in
  Computer Science}, pages 196--210. Springer Berlin Heidelberg, 2006.

\bibitem{russellnorvig}
S.~Russell and P.~Norvig.
\newblock {\em Artificial Intelligence: {A} Modern Approach}.
\newblock Prentice-Hall, 3rd edition, 2010.

\bibitem{Sciclu12}
N.~Scicluna, E.~Gatt, O.~Casha, I.~Grech, and J.~Micallef.
\newblock Fpga-based autonomous parking of a car-like robot using fuzzy logic
  control.
\newblock In {\em Electronics, Circuits and Systems (ICECS), 2012 19th IEEE
  International Conference on}, pages 229--232, Dec 2012.

\bibitem{Thr06}
S.~Thrun, W.~Burgard, and D.~Fox.
\newblock {\em Probabilistic Robotics}.
\newblock MIT Press, Cambridge, 2006.

\bibitem{Wan00}
E.~Wan and R.~Van~der Merwe.
\newblock {The Unscented Kalman Filter for Nonlinear Estimation}.
\newblock In {\em Adaptive Systems for Signal Processing, Communications, and
  Control Symposium 2000. AS-SPCC. The IEEE 2000}, pages 153--158, 2000.

\bibitem{Wang14}
Q.~Wang, P.~Zuliani, S.~Kong, S.~Gao, and E.~M. Clarke.
\newblock Sreach: Combining statistical tests and bounded model checking for
  nonlinear hybrid systems with parametric uncertainty.
\newblock {\em CoRR}, abs/1404.7206, 2014.

\end{thebibliography}
\balancecolumns
\end{document}